\pdfoutput=1

\documentclass[11pt]{article}
\usepackage{float}

\usepackage[]{acl}

\usepackage{times}
\usepackage{latexsym}
\usepackage{adjustbox}

\usepackage[T1]{fontenc}

\usepackage[utf8]{inputenc}

\usepackage{microtype}
\usepackage{graphicx}
\usepackage{adjustbox}

\usepackage{csvsimple}
\usepackage{array}
\usepackage{booktabs}
\title{Analyzing Bagging Methods for Language Models}

\author{Pranab Islam \\ pfi203@nyu.edu \\ New York University \And
         Shaan Khosla\\ shaan.khosla@nyu.edu \\ New York University \And     Arthur Lok \\ arthurmlok@nyu.edu \\New York University\And   Mudit Saxena \\ ms12768@nyu.edu \\ New York University}
\usepackage{booktabs}

\begin{document}
\maketitle
\begin{abstract}

Modern language models leverage increasingly large numbers of parameters to achieve performance on natural language understanding tasks. Ensembling these models in specific configurations for downstream tasks show even further performance improvements. In this paper, we perform an analysis of bagging language models and compare single language models to bagged ensembles that are roughly equivalent in terms of final model size. We explore an array of model bagging configurations for natural language understanding tasks with final ensemble sizes ranging from 300M parameters to 1.5B parameters and determine that our ensembling methods are at best roughly equivalent to single LM baselines. We note other positive effects of bagging and pruning in specific scenarios according to findings in our experiments such as variance reduction and minor performance improvements.

\end{abstract}

\section{Introduction}
\begin{figure*}
    \centering
    \includegraphics[width=12cm]{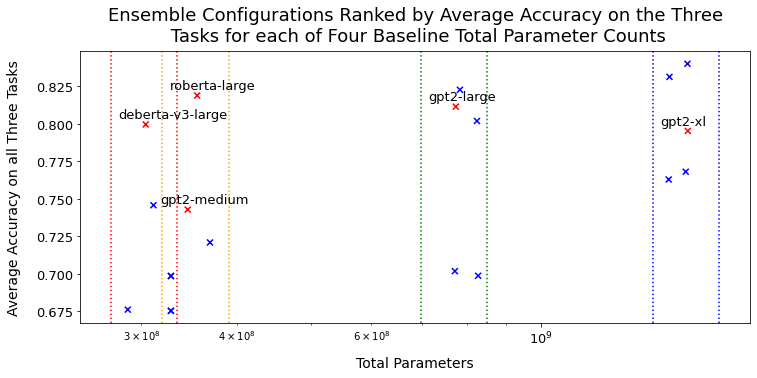}
    \caption{For all four baseline total parameter sizes (304M, 355M, 774M, and 1,558M), we show the performance of the five best models (which include both single LMs and ensembles). The color-coded dotted vertical lines represent intervals in which we consider models to be roughly equivalent in total number of parameters ($\pm 10\%$ of the four baseline total parameter sizes is considered roughly equivalent). Red crosses represent single LM performance; blue crosses are ensembles.}
    \label{fig:summary}
\end{figure*}
In this paper, we investigate the effects of ensembling language models in a number of different configurations, while keeping the ensemble's total parameter count roughly constant with respect to a single baseline language model, to study if there are any general or task-specific performance gains that can be established. In parallel, we study our methods to analyze the extent of other positive effects such as reducing the variance of model predictions.

The main techniques we leverage in our configurations are bootstrap aggregating (bagging) and pruning of language models. Bootstrap aggregation of large language models such as BERT \cite{bert} have clear performance gains on various downstream tasks \cite{bagging_aggression} \cite{rakotoson2021bagbert}. In the opposite direction in terms of model size, distilled versions of models such as BERT \cite{sanh2019distilbert} and pruned models \cite{michel2019sixteen} have performed comparably to the original model in experiments. We therefore employ pruning in some ensemble configurations in order to reduce the number of parameters per model, allowing us to create ensembles with more components while keeping total parameter amount constant.

Through our experiments, we provide a survey and comparison of different ensembling configurations against a single language model's performance on three of eight tasks in the SuperGLUE benchmark \cite{wang2019superglue}, allowing us to evaluate on a diverse set of difficult natural language tasks.


\section{Related Work}
Bootstrap aggregation of BERT has been explored in the field and has been shown to be effective at various downstream tasks. Prior work has shown various ensembles of BERT being used to solve a specific task such as online aggression identification \cite{bagging_aggression}, offensive Tweet classification \cite{offensive_tweet}, or multi-class topic classification on COVID-19 \cite{rakotoson2021bagbert}. 

Ensembles of other language models such as ALBERT and RoBERTa have shown promising results as well where research has show that ensembles of language models perform better than individual models on the Question Answering task \cite{ensembleroberta}. Furthermore, on the SuperGLUE leaderboard, an ensemble of RoBERTa is currently ranked 13$^{th}$. In addition to this work, others have shown that an ensemble of ALBERT was able to outperform large individual language models \cite{ensemblesquad}.

There is convincing evidence that performance of language models can be mostly maintained even if model sizes are reduced via distillation or pruning. DistilBERT manages to reduce BERT model size by 40$\%$ \cite{sanh2019distilbert} and improving inference times by 60$\%$ while maintaining close performance to BERT. Pruning has also shown promise with work from \citet{michel2019sixteen} showing that many of the attention heads in multi-headed attention are not necessary for maintaining model performance in machine translation and natural language inference. Interestingly, in some cases, removing attention heads could potentially improve performance according to \citet{michel2019sixteen}. 

In the literature, generally when ensembling models, the final models are much larger than the single language models that are used as the components of the ensembles. Furthermore, there typically is a specific task that the ensemble is designed to excel in. In contrast, we create size-efficient ensembles and test their generalized performance in addition to specific-task performance.

\section{Data}
The data used in our experiments and model development were all datasets that are included in the SuperGLUE benchmark. We focus on three of the eight total tasks in SuperGLUE, namely: Boolean Questions (BoolQ) \cite{clark2019boolq}, CommitmentBank (CB), and Recognizing Textual Entailment (RTE) \cite{wang2019superglue}. Combined, the three tasks allow us to test a broad range of natural language processing tasks including question-answering, textual entailment, and reading comprehension. All tasks included a train, validation, and test set. Due to the fact that the test set did not include labels, we repurpose 50$\%$ of the validation set to be a test set so that we can perform a final evaluation. Task performance was evaluated on either F1 or accuracy depending on the specific task.

\section{Models}

For our experiments, we will be using several different Transformer-based language models, \cite{transformers} including RoBERTa, DeBERTa, and GPT-2. Our selection is based on the desire to have a variety of model sizes and architectures represented in our analysis. Moreover, practitioners have access to these models and most of these models have shown excellent results on the SuperGLUE leaderboard.



\section{Methods}

\subsection{Bagging Methodology}
For a given baseline language model ($\emph{e.g.}$ GPT2-Large or DeBERTa-v3-Large), we fine-tune separate copies on the respective SuperGLUE task. We choose a hyperparameter search space based on generally recommended search spaces such as the one given by \cite{liu2019roberta} and then choose the hyperparameter set that performs best on the validation set based on an accepted metric for that dataset. For RTE and BoolQ, this is accuracy and for CB, this is Macro-F1.

We then use those hyperparameters and train $n$ models on $n$ bootstrapped versions of the training dataset. Depending on the ensemble configuration, we may prune these models before evaluation on the test set. Each model then produces a prediction of the probability of the input belonging to each class, and we conduct an equal-weighted soft majority vote for the final class choice. The predicted class is the one that has the highest sum of the $n$ models' predicted probability for that class.

In order to analyze the model prediction variance for single LMs compared to ensembles, we employ a method we call "double bootstrapping". We do so to fairly compare the variance of ensembles with single LMs. This method draws $m$ bootstrapped samples from each of the $n$ already-bootstrapped samples. Single LMs are trained on each double bootstrapped sample and ensembled according to the respective original bootstrapped sample they were drawn from. This results in a set of $n$ single LMs trained on bootstrapped samples and a set of $n$ ensembles, consisting of $m$ models each, trained on the double bootstrapped samples. Each of the $n$ ensembles and single LMs are thus exposed to the same original bootstrapped sample and a roughly fair comparison is made.
See Appendix \ref{appendix:b} for a visualization.
We analyze the variance of accuracy and F1 depending on the respective task for both sets of models and compare whether our bagging method reduces variance. The variance analysis was conducted on our large baseline models: RoBERTa-Large, DeBERTa-v3-Large, GPT2-Large, and GPT2-XL.

\subsubsection{Pruning Methodology}
In some ensemble configurations, we utilize pruning to decrease the size of models in the ensemble. This helps us create total ensemble sizes more comparable to the single LM while ideally maintaining model performance. Our pruning technique involves setting $X\%$ of parameters in a model with the lowest absolute values to zero. We choose $X$ based on the type of experiment we want to run.

\subsection{Experimental Design}

We ran over 100 configurations in our ensembles, using different combinations of model families, pruning, and bagging. Each configuration is evaluated on the test set on all three of our tasks. Depending on whether we use multiple models of the same type in a configuration, we use the version trained on the full data or multiple versions trained on bootstrapped samples. 

Many configurations were designed with ablation studies in mind and analyzed as sets of configurations. For example, configuration 16 is the same as configuration 300, with the exception of including a full RoBERTa-Large in the ensemble. Configurations 32, 301, and others were one-off deviations (model inclusions, prune vs non-prune, etc.) from configurations 16 and 300. Our analysis is heavily based on analyzing sets of similar configurations. 
We broadly classify our configurations into 6 types:

\textbf{Configuration Type 1: Single Model} uses a single model of a single model type ($\emph{e.g.}$ roberta-base). We evaluated all model types in our experiments in this configuration. In particular, DeBERTa-v3-Large, RoBERTa-Large, GPT2-Large, and GPT2-XL served as the basis of comparison to other configurations with a similar number of parameters.

\textbf{Configuration Types 2 and 3: Ensemble - Homogeneous Model Type With and Without Pruning} use multiple models of a single model type ($\emph{e.g.}$ two DeBERTa-Base). We often designed these configurations to compare ensembles of smaller models to larger models in the same family, holding number of parameters roughly equal in both. We also experimented with pruning in some ensembles, testing the effectiveness of different levels of pruning. One common set of configurations we conducted across all model families was creating ensembles of pruned models that have a similar number of parameters to the larger single model baseline. 

\textbf{Configuration Type 4: Ensemble - Heterogeneous Model Type - Same Model Family} use multiple model types within the same model family ($\emph{e.g.}$ GPT2-Medium and GPT2-Large) and allowed us to evaluate whether there were any gains from ensembling smaller versions of the larger models.

\textbf{Configuration Types 5 and 6: Ensemble - Heterogenous Model Type - Different Model Families With and Without Pruning} use model types across different model families ($\emph{e.g.}$ RoBERTa-Base and GPT2-Medium). These configurations aim to take advantage of the unique strengths of the different model families as well as benefit from different tokenization schemes and vocabularies.


\section{Results}

We report the top 15 configurations by average accuracy on BoolQ, CB, and RTE in Appendix \ref{appendix:a}. For all configurations tested, please check \path{Experiment Results.csv} in our code repo. 

Through our experiments, we conclude that applying bagging with equal-weighted soft majority vote and pruning techniques to create ensembles does not outperform single language models of roughly equal size in terms of overall performance on these three tasks (see Figure \ref{fig:summary}). We see some task-specific gains in some ensembles, and even some of our top configurations across the board were ensembles of models in different model families. However, at the parameter ranges of our main baseline models, single model baselines are consistently in the top three performing models for each parameter count size, outperforming the majority of ensembles constructed.

We note that pruning models of more than 20\% of their weights tend to yield extremely poor results, though recogniz either negligible or even positive effects at small amounts of pruning which is discussed in Section \ref{section:discussion}. We also see task-specific and model-specific gains using bagging in our results. This is corroborated by our variance analysis which showed positive effects of variance reduction and in some cases on accuracy when comparing our single bootstrapped model to our double bootstrapped models. RoBERTa-Large in particular had a average reduction of 5.2\% in the standard deviation of its prediction accuracy across the tasks along with a 4.9\% average accuracy improvement.

\section{Discussion}
\label{section:discussion}

Despite our ensembles not showing definitive performance gains compared to the large single LM baselines, we note the general positive effects of our methods which may be useful for practitioners and identify paths for further experimentation that could be promising based on our findings. 

With regard to bagging, the single language model was rarely outperformed by ensembles of any kind in the same size. However, our bagging method did improve performance in cases where there was evidence of overfitting and high variance, particularly with specific tasks. Two of our models, RoBERTa-Base and DeBERTa-Base, performed poorly on the RTE and CB test set, scoring close to random chance despite doing well on the validation set. In ensembles with the 3-5 models of the respective model (\emph{e.g.} configuration 55 versus configuration 67 and 68 for DeBERTa) trained on bootstrapped samples, we saw minor improvements in both tasks. There were multiple cases where the smaller or distilled version of models actually performed better on certain tasks than the larger ones which shows further evidence that bagging may help in cases where large models can overfit. 

The variance analysis conducted on our large baseline models also supports this conclusion (see Appendix \ref{appendix:c} and Appendix \ref{appendix:d}). The single bootstrapped models have an average standard deviation of 4.7\% in their accuracy compared to the double bootstrapped models' 3.5\%. Out of the 12 model/task combinations that were tested, 6 combinations improved in accuracy. Overall, these results point to the fact that relatively small differences in training data can severely affect performance in large language models, and bagging is a useful tool for practitioners to combat this in their research. 

Results of our simple pruning approach provided further support for conclusions in past work: large language models do not necessarily need all their parameters in order to still perform well or even improve. We noticed that for DistilRoBERTa, RoBERTa, DistilGPT2, GPT2, and GPT2-medium that pruning up to 5\% of the model parameters consistently made the final LM average accuracy on the three tasks roughly equal or slightly better than the non-pruned single LM. Some specific examples include our 27$^{th}$ configuration, which is a 5\% pruned ensemble of GPT2-medium, RoBERTa-large, and RoBERTa-base. Configuration 27 had a notable 2.1\% superior average performance on the three tasks than the non-pruned configuration. This is supporting evidence that practitioners ought to consider lightly pruning models and ensembles of models after fine-tuning (unless heavy dropout and regularization is already used) in order to reduce model overhead and inference time.

In future work, we would like to use these findings to improve our methodologies in different ways. Given that the best models for each task were from different model families and the best models in terms of overall average accuracy were multi-model family ensembles, a boosting ensembling method may be a more effective method to leverage the strengths of a diverse set of models. Also, our current method weights all models in the ensemble equally, but could benefit from a more dynamic assignment that takes into account each member's effectiveness on a task.

Additionally, we see that pruning at lower levels could still be useful. Instead of pruning after the fine-tuning procedure is completed, it would likely be beneficial to prune iteratively. We notice that large parts of the model are unused, and pruning them while continuing to train the model may improve the overall effects of pruning.

\section{Ethical Considerations}

We have shown that bagging is useful in some specific applications for practitioners to use. However, we encourage practitioners to not blindly increase the number of models, and thus the number of parameters they are training, as this has been shown in other papers to have negative environmental effects \cite{environment}. Similarly, although we did not see many benefits of pruning, we encourage further investigation in to the use of pruning, as this may counteract the negative environmental effects of larger language models \cite{environment}. 

We also highlight some risks of bagging several large language models with pruning. In this paper, we did not analyze the effects of bagging on harmful biases resulting from stereotyping that propagate negative generalizations involving gender, race, religion, and other social constructs \cite{bias}. We encourage practitioners that bag large language models to do bias checks and proper model validation before deployment.

\section{Github Repo and Experiment Tracker}

The code for our project includes scripts for the end-to-end process of running our experiments including procuring SuperGLUE data, fine-tuning models on the full validation or bagged samples, and evaluating configurations on the test set. The repo can be found at: \url{https://github.com/shaankhosla/AggregateLMs}. We also created an experiment tracker which contains detailed results and documentation of our configurations and variance analysis experiments which can be found here: \url{https://docs.google.com/spreadsheets/d/1nVZOPeP8s_zMcnDBw9QRAu_UI3jdbICFpolVc4rwT9A/edit?usp=sharing}.


\clearpage

\bibliography{analyzing_bagging_methods_for_language_models}

\begin{thebibliography}{14}
\expandafter\ifx\csname natexlab\endcsname\relax\def\natexlab#1{#1}\fi

\bibitem[{Bachina et~al.(2021)Bachina, Balumuri, and Kamath}]{ensembleroberta}
Sony Bachina, Spandana Balumuri, and Sowmya Kamath. 2021.
\newblock Ensemble albert and roberta for span prediction in question
  answering.
\newblock In \emph{Proceedings of the 1st Workshop on Document-grounded
  Dialogue and Conversational Question Answering (DialDoc 2021)}, pages 63--68.

\bibitem[{Clark et~al.(2019)Clark, Lee, Chang, Kwiatkowski, Collins, and
  Toutanova}]{clark2019boolq}
Christopher Clark, Kenton Lee, Ming-Wei Chang, Tom Kwiatkowski, Michael
  Collins, and Kristina Toutanova. 2019.
\newblock Boolq: Exploring the surprising difficulty of natural yes/no
  questions.
\newblock In \emph{NAACL}.

\bibitem[{Devlin et~al.(2018)Devlin, Chang, Lee, and Toutanova}]{bert}
Jacob Devlin, Ming{-}Wei Chang, Kenton Lee, and Kristina Toutanova. 2018.
\newblock \href {http://arxiv.org/abs/1810.04805} {{BERT:} pre-training of deep
  bidirectional transformers for language understanding}.
\newblock \emph{CoRR}, abs/1810.04805.

\bibitem[{Huang et~al.(2022)Huang, Li, Chen, Claesen, Xi, Chen, Jiang, Liu,
  Xiong, and Yan}]{environment}
Kai Huang, Bowen Li, Siang Chen, Luc Claesen, Wei Xi, Junjian Chen, Xiaowen
  Jiang, Zhili Liu, Dongliang Xiong, and Xiaolang Yan. 2022.
\newblock Structured term pruning for computational efficient neural networks
  inference.
\newblock \emph{IEEE Transactions on Computer-Aided Design of Integrated
  Circuits and Systems}.

\bibitem[{Li et~al.(2017)Li, Li, Peng, Shazeer, Parmar, Uszkoreit, Jones,
  Gomez, Kaiser, and Polosukhin}]{transformers}
Shilun Li, Renee Li, Ashish Peng, Veronica~Vaswani, Noam Shazeer, Niki Parmar,
  Jakob Uszkoreit, Llion Jones, Aidan~N. Gomez, Lukasz Kaiser, and Illia
  Polosukhin. 2017.
\newblock Attention is all you need.
\newblock \emph{arXiv preprint arXiv:1706.03762}.

\bibitem[{Li et~al.(2021)Li, Li, and Peng}]{ensemblesquad}
Shilun Li, Renee Li, and Veronica Peng. 2021.
\newblock Ensemble albert on squad 2.0.
\newblock \emph{arXiv preprint arXiv:2110.09665}.

\bibitem[{Liang et~al.(2021)Liang, Wu, Morency, and Salakhutdinov}]{bias}
Paul~Pu Liang, Chiyu Wu, Louis-Philippe Morency, and Ruslan Salakhutdinov.
  2021.
\newblock Towards understanding and mitigating social biases in language
  models.
\newblock In \emph{International Conference on Machine Learning}, pages
  6565--6576. PMLR.

\bibitem[{Liu et~al.(2019)Liu, Ott, Goyal, Du, Joshi, Chen, Levy, Lewis,
  Zettlemoyer, and Stoyanov}]{liu2019roberta}
Yinhan Liu, Myle Ott, Naman Goyal, Jingfei Du, Mandar Joshi, Danqi Chen, Omer
  Levy, Mike Lewis, Luke Zettlemoyer, and Veselin Stoyanov. 2019.
\newblock Roberta: A robustly optimized bert pretraining approach.
\newblock \emph{arXiv preprint arXiv:1907.11692}.

\bibitem[{Michel et~al.(2019)Michel, Levy, and Neubig}]{michel2019sixteen}
Paul Michel, Omer Levy, and Graham Neubig. 2019.
\newblock Are sixteen heads really better than one?
\newblock \emph{Advances in neural information processing systems}, 32.

\bibitem[{Nikolov and Radivchev(2019)}]{offensive_tweet}
Alex Nikolov and Victor Radivchev. 2019.
\newblock Nikolov-radivchev at semeval-2019 task 6: Offensive tweet
  classification with bert and ensembles.
\newblock In \emph{Proceedings of the 13th international workshop on semantic
  evaluation}, pages 691--695.

\bibitem[{Rakotoson et~al.(2021)Rakotoson, Letaillieur, Massip, and
  Laleye}]{rakotoson2021bagbert}
Lo{\"\i}c Rakotoson, Charles Letaillieur, Sylvain Massip, and Fr{\'e}jus
  Laleye. 2021.
\newblock Bagbert: Bert-based bagging-stacking for multi-topic classification.
\newblock \emph{arXiv preprint arXiv:2111.05808}.

\bibitem[{Risch and Krestel(2020)}]{bagging_aggression}
Julian Risch and Ralf Krestel. 2020.
\newblock Bagging bert models for robust aggression identification.
\newblock In \emph{Proceedings of the Second Workshop on Trolling, Aggression
  and Cyberbullying}, pages 55--61.

\bibitem[{Sanh et~al.(2019)Sanh, Debut, Chaumond, and
  Wolf}]{sanh2019distilbert}
Victor Sanh, Lysandre Debut, Julien Chaumond, and Thomas Wolf. 2019.
\newblock Distilbert, a distilled version of bert: smaller, faster, cheaper and
  lighter.
\newblock \emph{arXiv preprint arXiv:1910.01108}.

\bibitem[{Wang et~al.(2019)Wang, Pruksachatkun, Nangia, Singh, Michael, Hill,
  Levy, and Bowman}]{wang2019superglue}
Alex Wang, Yada Pruksachatkun, Nikita Nangia, Amanpreet Singh, Julian Michael,
  Felix Hill, Omer Levy, and Samuel Bowman. 2019.
\newblock Superglue: A stickier benchmark for general-purpose language
  understanding systems.
\newblock \emph{Advances in neural information processing systems}, 32.

\end{thebibliography}
\bibliographystyle{acl_natbib}

\onecolumn
\appendix
\label{appendix}
\clearpage

\section{Appendix: Top 15 Configurations by Average Accuracy}
\label{appendix:a}
    
\begin{table}[h]
\footnotesize
    \centering
    \begin{tabular}{|>{\centering}p{0.05\textwidth}|p{0.05\textwidth}|p{0.05\textwidth}|p{0.05\textwidth}|p{0.05\textwidth}|p{0.05\textwidth}|p{0.225\textwidth}|p{0.225\textwidth}|p{0.06\textwidth}|}
    \hline
        Config & BoolQ Acc. & CB Acc. & CB Macro-F1 & RTE Acc. & Avg. Acc. & Experiment Type & List Models & Total Params (M) \\ \hline
        32 & 79.33\% & 92.86\% & 91.19\% & 79.86\% & 84.02\% & Ensemble - Heterogeneous Model Type - Different Model Families & gpt2-large, gpt2-medium, roberta-large, distilroberta-base & 1556 \\ \hline
        16 & 78.87\% & 92.86\% & 91.19\% & 77.70\% & 83.14\% & Ensemble - Heterogeneous Model Type - Different Model Families & gpt2-large, gpt2-medium, roberta-large & 1474 \\ \hline
        27 & 80.55\% & 85.71\% & 88.89\% & 80.58\% & 82.28\% & Ensemble - Heterogeneous Model Type - Different Model Families (Pruned Models) & gpt2-medium, roberta-base, roberta-large & 784 \\ \hline
        8 & 82.32\% & 92.86\% & 89.58\% & 70.50\% & 81.89\% & Single Model & roberta-large & 355 \\ \hline
        21 & 75.78\% & 92.86\% & 94.21\% & 74.82\% & 81.15\% & Single Model & gpt2-large & 774 \\ \hline
        26 & 80.06\% & 85.71\% & 88.89\% & 74.82\% & 80.20\% & Ensemble - Heterogeneous Model Type - Different Model Families & gpt2-medium, roberta-base, roberta-large & 825 \\ \hline
        58 & 87.83\% & 60.71\% & 42.72\% & 91.37\% & 79.97\% & Single Model & deberta-v3-large & 304 \\ \hline
        20 & 78.04\% & 82.14\% & 86.32\% & 78.42\% & 79.53\% & Single Model & gpt2-xl & 1558 \\ \hline
        12 & 72.78\% & 83.93\% & 87.64\% & 73.74\% & 76.82\% & Ensemble - Homogeneous Model Type & gpt2-large & 1548 \\ \hline
        15 & 70.52\% & 85.71\% & 75.71\% & 72.66\% & 76.30\% & Ensemble - Homogeneous Model Type (Pruned Models) & gpt2-large & 1471 \\ \hline
        17 & 81.83\% & 67.86\% & 47.90\% & 76.26\% & 75.32\% & Ensemble - Heterogeneous Model Type - Different Model Families & deberta-v3-large, gpt2-medium, roberta-large, deberta-base, gpt2, deberta-v3-small, deberta-v3-xsmall, distilroberta-base, roberta-base, distilgpt2 & 1562 \\ \hline
        40 & 73.39\% & 82.14\% & 72.69\% & 69.06\% & 74.86\% & Ensemble - Homogeneous Model Type (Pruned Models) & gpt2-large & 1424 \\ \hline
        81 & 74.68\% & 78.57\% & 55.23\% & 70.50\% & 74.58\% & Ensemble - Heterogeneous Model Type - Different Model Families (Pruned Models) & gpt2-medium, deberta-v3-small & 311 \\ \hline
        54 & 70.95\% & 82.14\% & 80.88\% & 69.78\% & 74.29\% & Single Model & gpt2-medium & 345 \\ \hline
        104 & 70.52\% & 78.57\% & 70.42\% & 72.66\% & 73.92\% & Ensemble - Homogeneous Model Type (Pruned Models) & gpt2-large & 1084 \\ \hline
    \end{tabular} 
    \caption{Top 15 Configurations by Average Accuracy. Table also reports accuracy for each task, the list of models used in the ensemble, and total parameters in the ensemble. Full results can be found on our code repo \url{https://github.com/shaankhosla/AggregateLMs}.}
\label{tab:top30configs}
\end{table}




\clearpage
\section{Appendix: Double Bootstrapping Method}
\label{appendix:b}

\begin{figure}[h]
    \centering
    \includegraphics[width=16cm]{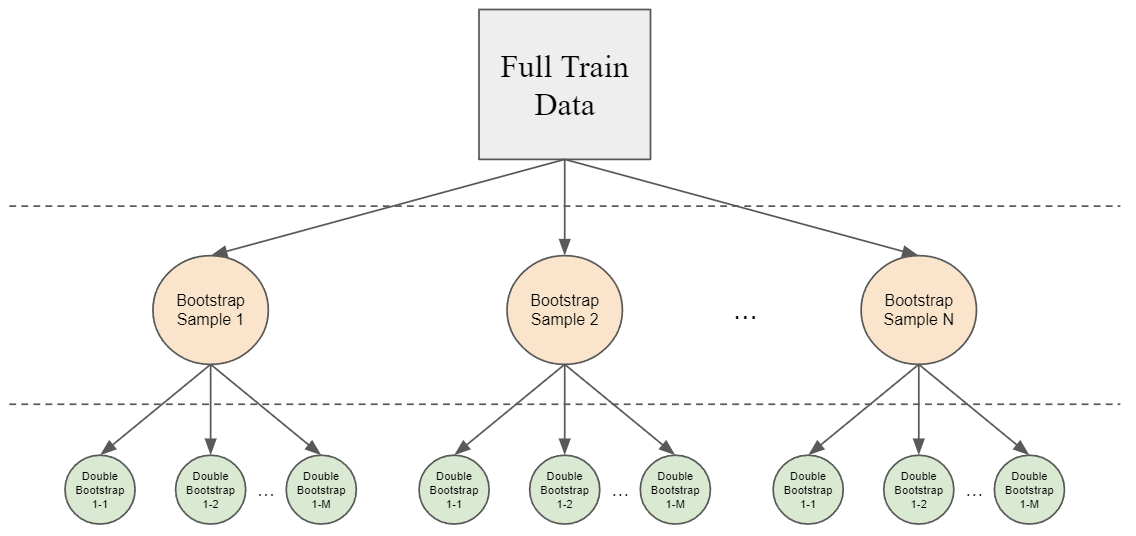}
    \caption{Double bootstrapping method. The diagram above shows the double bootstrapping method used in our variance analysis. Models are fit to each double bootstrap sample and are then ensembled with respect to each bootstrap sample that they were bootstrapped from. These ensembles are compared to models fit on the first level of bootstrap samples.}
    \label{fig:double_bootstrap}
\end{figure}

\section{Appendix: Average Accuracy of Single Bootstrapped and Double Bootstrapped Models}
\label{appendix:c}

\begin{table*}[!htbp]
    \centering
    \begin{tabular}{cccc}%
    \bfseries Model & \bfseries Task & \bfseries Single Bootstrapped Performance & \bfseries Double Bootstrapped Performance 
    \begin{filecontents*}{full_variance_analysis.csv}
model name,task,single_accuracy_avg,double_bootstrapped_accuracy_avg
gpt-large,BoolQ,70\%,69\%
gpt-large,CB,80\%/\77%,73\%/66\%
gpt-large,RTE,72\%,71\%
gpt2-xl,BoolQ,71\%,69\%
gpt2-xl,CB,74\%/68\%,70\%/64\%
gpt2-xl,RTE,71\%,74\%
roberta-large,BoolQ,79\%,80\%
roberta-large,CB,71\%/72\%,79\%/76\%
roberta-large,RTE,65\%,71\%
deberta-large,BoolQ,60\%,62\%
deberta-large,CB,60\%/30\%,40\%/20\%
deberta-large,RTE,51\%,52\%
\end{filecontents*}
\csvreader[head to column names]{full_variance_analysis.csv}{}
    {\\\hline\csvcoli&\csvcolii&\csvcoliii&\csvcoliv}
    \end{tabular}
    \caption{Performance of single bootstrapped and double bootstrapped models. The table above shows average accuracy of the single and ensemble of double bootstrapped models in each task. For CB, we show an Accuracy/Macro-F1 pair. In several scenarios, accuracy increases on the task when using the ensemble.}

\end{table*}


\clearpage
\section{Appendix: Standard Deviation of Accuracy of Single Bootstrapped and Double Bootstrapped Models}
\label{appendix:d}

\begin{table*}[!htbp]
\centering
\begin{tabular}{c|ccc|ccc|ccc}
\toprule
Model &  \multicolumn{3}{c}{CB} & \multicolumn{3}{c}{BoolQ}& \multicolumn{3}{c}{RTE}\\
\midrule
{}   & Single   & Double    & Diff   & Single   & Double    & Diff & Single   & Double    & Diff\\
GPT-Large   &  3.4 & 4.5   & 1.1  & 1.3 & 0.8 & -0.5 & 2.4 & 2.4 & 0\\
GPT2-XL   &  5.3 & 4.6   & -0.7  & 1.4 & 0.8 & -0.6 & 8.4 & 2.2 & -6.2\\
RoBERTa Large   & 7.3  & 3.7   & -3.6  & 6.4 & 1.3 & -5.1 & 11 & 4.4 & -6.6\\
DeBERTa Large   & 4.5  & 13.8   & 9.3  & 1.2 & 0.3 & -0.9 & 3.5 & 2.9 & -0.7\\
\bottomrule
\end{tabular}
\caption{Standard deviation analysis of single bootstrapped and double bootstrapped models. All units are in \% accuracy. In nearly all scenarios, standard deviation decreases.}
\end{table*}

\end{document}